\documentclass{article}

\usepackage{PRIMEarxiv}

\usepackage[utf8]{inputenc} % allow utf-8 input
\usepackage[T1]{fontenc}    % use 8-bit T1 fonts
\usepackage{hyperref}       % hyperlinks
\usepackage{url}            % simple URL typesetting
\usepackage{booktabs}       % professional-quality tables
\usepackage{amsmath}      % blackboard math symbols
\usepackage{nicefrac}       % compact symbols for 1/2, etc.
\usepackage{microtype}      % microtypography
\usepackage{lipsum}
\usepackage{fancyhdr}       % header
\usepackage{graphicx}       % graphics
\usepackage{subcaption}
\usepackage{adjustbox}
\usepackage{enumitem} % 列表支持
\usepackage{booktabs}

\graphicspath{{media/}}     % organize your images and other figures under media/ folder

\title{Grid-Augmented Vision: A Simple yet Effective Approach for Enhanced Spatial Understanding in Multi-Modal Agents
\thanks{\textit{\underline{Citation}}: 
\textbf{Authors. Title. Pages.... DOI:000000/11111.}} 
} % 修改脚注编号

\usepackage{authblk}
\author[1]{\textbf{JOONGWON CHAE\#}}
\author[1]{\textbf{Zhenyu Wang\#}}
\author[3]{\textbf{Lian Zhang*}}
\author[2]{\textbf{Dongmei Yu*}}
\author[1]{\textbf{Peiwu Qin*}}
\affil[1]{Institute of Biopharmaceutical and Health Engineering, Shenzhen International Graduate School, Tsinghua University, Shenzhen, Guangdong, China}
\affil[2]{School of Mechanical, Electrical \& Information Engineering, Shandong University, Weihai, Shandong, China}
\affil[3]{The First Hospital of Hebei Medical University, Shijiazhuang, Hebei, China}
\affil[ \#]{These authors contributed equally to this work.}
\affil[*]{Corresponding authors: \texttt{pwqin@sz.tsinghua.edu.cn} (P. Qin)}

\begin{document}
\maketitle

\begin{abstract}
	
Recent advances in multimodal models have demonstrated impressive capabilities in object recognition and scene understanding. However, these models often struggle with precise spatial localization - a critical capability for real-world applications. We propose a simple yet effective backbone-free
approach that introduces an explicit visual reference system for enhanced spatial understanding. Byoverlaying a 9×9 black grid pattern onto input images, our method provides consistent spatial anchors without requiring architectural modifications or additional computational overhead. Experiments on the COCO 2017 dataset demonstrate that our grid-based approach achieves significant improvements in localization accuracy, with a 107.4\% increase in IoU (from 0.27 to 0.56) and a 194.4\% improvement in GIoU (from 0.18 to 0.53) compared to baseline performance. These results highlight how explicit spatial references can effectively bridge the gap between conceptual understanding and precise localization. Our method’s simplicity and effectiveness make it particularly valuable for applications requiring accurate spatial reasoning, such as robotic manipulation, medical imaging, and autonomous navigation. The project code is available at \url{https://github.com/triumph123aaa/GRID-AUGMENTED-VISION} .

\end{abstract}


\section{Introduction}
Recent advances in multimodal models have revolutionized computer vision and natural language processing~\cite{alayrac2022flamingo,lai2024lisa,liu2024visual}, enabling sophisticated understanding of visual content through language interactions. These models, particularly CLIP~\cite{radford2021learning}, have demonstrated remarkable capabilities in connecting visual and linguistic information, achieving impressive zero-shot performance across various tasks. However, despite these advances, these models face a significant
challenge in precise spatial localization—a critical capability required for bridging conceptual understanding with real-world applications. For instance, while these models excel at identifying objects ("there is a cat in the image"), they often struggle with precise spatial descriptions ("the cat is located at coordinates (342, 156)"). This spatial localization capability is crucial for applications such as robotic manipulation~\cite{zeng2018learning}, medical image analysis~\cite{ronneberger2015u}, and autonomous navigation~\cite{belanger2017end}. 

The challenge of spatial understanding in vision models has been approached through various methods. Traditional approaches rely heavily on vision backbones (e.g., ResNet~\cite{he2016deep}, DenseNet~\cite{huang2017densely}) to extract visual features. While these backbone-based methods effectively recognize objects, they struggle with precise coordinate mapping~\cite{yang2020rethinking}, primarily because the spatial information becomes increasingly abstract through successive layers of processing. The emergence of Vision Transformer (ViT)~\cite{dosovitskiy2020image} introduced a new paradigm, treating images as sequences of patches and utilizing positional encodings to maintain spatial information. However, these positional encodings, typically implemented as learned embeddings or sinusoidal functions, provide only implicit spatial information, making precise coordinate prediction challenging.

A key insight from transformer architectures is the critical role of positional encoding in spatial understanding. In language transformers, positional encoding enables the model to understand word order and relative positions. Similarly, in ViT, positional encodings help maintain the spatial relationship between image patches. However, current approaches to positional encoding in vision transformers remain abstract and indirect—they are numerical embeddings that the model must learn to interpret, rather than explicit spatial references that directly guide spatial understanding.

Our work is motivated by a fundamental question: Can we make positional encoding more explicit and visual? Drawing inspiration from how humans use grid-based references (like chess boards or map coordinates) for precise spatial
communication, we propose a novel approach to spatial understanding in multimodal models. Instead of relying on implicit numerical positional encodings, we introduce a visual grid system that serves as an explicit, visual form of positional encoding. By overlaying a 9 × 9 black grid pattern onto input images, we provide direct spatial reference points that the model can use for precise localization.

\section{Related Work}\label{sec:related_work}

\subsection{Position Information in Vision-Language Models}

Modern vision-language models have revolutionized our ability to understand and reason about visual content. At their core, these models must address a fundamental challenge: how to effectively encode and utilize spatial information. We begin by examining how different approaches handle this crucial aspect.

\subsubsection{Positional Encoding in Transformers}

The transformer architecture~\cite{vaswani2017attention} introduced positional encoding as a way to inject order information into sequence modeling. In the original transformer, this took the form of sinusoidal functions:

\begin{equation}
	PE_{(pos,2i)} = sin(pos/10000^{2i/d_{model}})
\end{equation}

\begin{equation}
	PE_{(pos,2i+1)} = cos(pos/10000^{2i/d_{model}})
\end{equation}

where $pos$ is the position and $i$ is the dimension. This approach proved crucial for language understanding, as it enabled the self-attention mechanism to consider token positions.

\subsubsection{Vision Transformer's Spatial Understanding}

Vision Transformer (ViT)~\cite{dosovitskiy2020image} adapted this concept to images by treating them as sequences of patches. However, the challenge in vision is fundamentally different from text: spatial relationships are two-dimensional and more complex. ViT handles this by:

\begin{itemize}
	\item Dividing images into fixed-size patches (typically 16×16 pixels)
	\item Adding learned position embeddings to patch embeddings
	\item Using self-attention to model relationships between patches
\end{itemize}

While effective for many tasks, this approach still struggles with precise spatial localization, as the position information becomes increasingly abstract through the layers.

\subsubsection{CLIP's Cross-Modal Position Understanding}

CLIP further extended this to cross-modal understanding, demonstrating impressive zero-shot capabilities. However, its handling of spatial information faces several challenges:

\begin{itemize}
	\item The position encoding is optimized for image-text matching rather than precise localization
	\item Spatial relationships are learned implicitly through natural language supervision
	\item The model lacks explicit mechanisms for coordinate prediction
\end{itemize}

These limitations become particularly apparent in tasks requiring precise spatial localization.

\subsection{Approaches to Spatial Understanding in Computer Vision}

The challenge of spatial understanding in computer vision has been addressed through various approaches, each with distinct characteristics and limitations.

\subsubsection{Traditional Vision Backbone Approaches}

Convolutional Neural Networks (CNNs) have traditionally handled spatial information through their architectural design. Networks like ResNet~\cite{he2016deep} and DenseNet~\cite{huang2017densely} process spatial information through hierarchical feature extraction, where skip connections and dense connectivity patterns help preserve spatial information throughout the network. The fundamental operation in these networks is convolution, which captures local spatial relationships:

\begin{equation}
	f_{spatial}(x) = CNN(x) \rightarrow abstract\ features
\end{equation}

However, this approach faces a fundamental limitation: as information flows through the network layers, precise spatial information becomes increasingly abstracted. While this hierarchical processing is effective for object recognition and scene understanding, it makes exact coordinate prediction particularly challenging. The spatial information becomes more semantic and less geometric, making it difficult to map back to precise coordinates in the original image space.

\subsubsection{Attention-Based Spatial Understanding}

Modern approaches have leveraged attention mechanisms for spatial understanding, introducing new ways to process and maintain spatial information. The core of these approaches is the self-attention mechanism:

\begin{equation}
	Attention(Q, K, V) = softmax(\frac{QK^T}{\sqrt{d_k}})V
\end{equation}

where $Q$, $K$, and $V$ represent queries, keys, and values derived from spatial features. While this mechanism allows for global spatial reasoning and has shown impressive results in various vision tasks, it introduces its own set of challenges. The computational complexity scales quadratically with spatial dimensions, making it impractical for high-resolution images. Moreover, the attention maps provide soft assignments rather than precise coordinates, and spatial information is still represented implicitly.

\subsubsection{Cross-Modal Spatial Understanding}

In multimodal systems, cross-attention mechanisms are employed to align spatial and linguistic information. These systems attempt to bridge the gap between textual descriptions and visual spatial information through learned attention patterns. Visual features attend to textual descriptions while language tokens attend to spatial regions, creating a bidirectional flow of information. The probability of a location given text and image can be expressed as:

\begin{equation}
	p(location|text,image) = softmax(f_{cross-attention}(text, image\ features))
\end{equation}

Despite the sophistication of these approaches, they still struggle with precise localization. The resulting probability distributions over locations often lack the precision needed for exact coordinate prediction, especially in real-world applications requiring precise spatial understanding.

\subsubsection{Direct Coordinate Prediction Approaches}

Recent work has attempted to address these limitations through direct coordinate prediction. Regression-based methods directly predict spatial coordinates through:

\begin{equation}
	(x, y) = f_{regression}(image\ features)
\end{equation}

However, these approaches often suffer from training instability and scale sensitivity. More sophisticated approaches like DETR~\cite{carion2020end} use learned object queries and direct set prediction of bounding boxes, enabling end-to-end training without traditional anchor boxes. While these methods show promise, they still face significant challenges in convergence speed, small object detection, and achieving the level of precision required for exact coordinate prediction in practical applications.

\subsection{Visual Position Encoding and Grid-Based Approaches}

While positional encoding has been extensively studied in transformer architectures, the concept of making this information visually explicit represents a relatively unexplored direction. This section examines relevant approaches that inform our visual position encoding method.

\subsubsection{Position Encoding Evolution in Vision Models}

The evolution of position encoding in vision models reveals an interesting trajectory. Traditional CNNs implicitly encode position through spatial convolutions and pooling operations. Vision Transformer (ViT) made this more explicit with learnable position embeddings for image patches, yet these remain in the abstract feature space. The fundamental challenge lies in bridging the gap between these abstract position representations and the need for precise spatial localization:

\begin{equation}
	z_{patch} = f_{embedding}(x_{patch}) + p_{learned}
\end{equation}

where $z_{patch}$ represents the final patch embedding, $f_{embedding}$ is the patch embedding function, and $p_{learned}$ is the learned position embedding. This approach, while effective for general vision tasks, still maintains position information in an implicit form.

\subsubsection{Grid-Based References in Computer Vision}

Grid-based approaches have appeared in various computer vision contexts, though with different objectives than our proposed method. In medical image analysis~\cite{ronneberger2015u}, regular grid patterns have been used for registration and alignment. Similarly, in autonomous navigation~\cite{chen2017end}, grid-like representations help in spatial mapping. However, these applications typically use grids as a data structure or processing tool, rather than as an explicit visual position encoding mechanism.

The concept of visual spatial references has historical precedents in human-designed systems. Chess boards, map coordinates, and architectural blueprints all demonstrate the effectiveness of grid-based spatial reference systems in human spatial reasoning. These systems succeed because they provide:

\begin{equation}
	location = f_{grid}(row, column) \rightarrow explicit\ spatial\ reference
\end{equation}

This direct mapping between grid coordinates and spatial locations offers a level of precision and interpretability that abstract position encodings struggle to achieve.

\subsubsection{Spatial Understanding in Multimodal Systems}

Recent work in multimodal systems has highlighted the importance of explicit spatial representations for cross-modal understanding. CLIP and similar models learn to align visual and textual representations, but often struggle with precise spatial references. The spatial understanding in these systems can be characterized as:

\begin{equation}
	s_{alignment} = f_{multimodal}(v_{visual}, t_{text}, p_{implicit})
\end{equation}

where $p_{implicit}$ represents implicit position information. This formulation reveals a key limitation: the lack of explicit spatial reference points that both modalities can easily ground to.

\subsection{Prompt Engineering and Visual Guidance}

The emergence of prompt engineering in vision-language models offers interesting parallels to our work. Prompts, functioning as instructional directives or conditional parameters supplied to artificial intelligence models, can manifest in either textual or visual modalities. Similar to how textual prompts guide language models, visual prompts can provide explicit guidance for spatial understanding. SAM~\cite{kirillov2023segment} demonstrated the power of visual prompts for segmentation tasks, though it focuses on region selection rather than coordinate prediction.

The relationship between prompts and spatial understanding can be expressed as:

\begin{equation}
	r_{spatial} = f_{model}(image, prompt_{visual}, prompt_{text})
\end{equation}

This formulation suggests the potential for explicit visual guidance in improving spatial understanding, a concept that our grid-based approach builds upon and extends.

Our work bridges these various threads of research by introducing a visual position encoding system that makes spatial references explicit and directly interpretable, while maintaining the advantages of modern deep learning approaches.
This combination addresses the limitations of current methods while opening new possibilities for precise spatial
understanding in multimodal systems.

\section{Method}\label{sec:methodology}

\subsection{Overview}
In this section, we present our simple yet effective approach to enhance multimodal models' spatial localization capabilities through a grid-based reference system. The core of our method lies in overlaying a $9 \times 9$ black grid onto input images, which serves as an explicit spatial reference without requiring traditional vision backbones.

\subsection{Grid Reference System Design}
Our grid reference system is intentionally designed to be straightforward and effective. The system employs a $9 \times 9$ black grid overlay on input images, with carefully chosen parameters to maximize effectiveness while minimizing interference with image content. The grid lines are set to black with an alpha transparency of $0.3$, striking a balance between visibility and non-intrusiveness. This transparency level ensures that the grid provides clear spatial references while preserving the integrity of the original image information.

The grid integration process follows a simple compositing function:
\begin{equation}
    I_g = \alpha \cdot G + (1 - \alpha) \cdot I,
\end{equation}
where $I_g$ represents the grid-enhanced image, $G$ is the grid pattern, $I$ is the original image, and $\alpha$ is the transparency factor ($0.3$). This straightforward approach ensures that both the grid reference and original image content remain clearly visible while maintaining the model's ability to process the combined information effectively.

\subsection{Implementation}
The implementation of our approach focuses on simplicity and effectiveness. The grid generation process is implemented through a straightforward algorithm that creates equally spaced horizontal and vertical lines across the image. The grid is generated once for each input image and can be easily scaled to different image dimensions while maintaining the $9 \times 9$ structure.

Key implementation details include:
\begin{itemize}
    \item \textbf{Grid resolution:} $9 \times 9$ fixed structure
    \item \textbf{Line color:} Black (\#000000)
    \item \textbf{Line transparency:} $0.3$ alpha value
    \item \textbf{Line width:} $1$ pixel
    \item \textbf{Integration method:} Simple alpha blending
\end{itemize}

These design choices were derived from extensive experiments to ensure optimal performance in terms of balance between visibility and minimal interference with the original image content.

As shown in Figure~\ref{fig:grid_process}, our grid overlay process transforms the original image (Figure~\ref{fig:grid_process}(a)) into an enhanced version with a uniformly distributed $9 \times 9$ black grid pattern (Figure~\ref{fig:grid_process}(b)), providing explicit spatial references while maintaining the clarity of the original image content.

\begin{figure}[h!]
	\centering
	\begin{subfigure}[b]{0.48\linewidth}
		\centering
		\includegraphics[width=\linewidth]{figures/000000190756.jpg}
		\caption{Original input image}
		\label{fig:grid_original}
	\end{subfigure}
	\hfill
	\begin{subfigure}[b]{0.48\linewidth}
		\centering
		\includegraphics[width=\linewidth]{figures/0000001907562.jpg}
		\caption{Image with grid overlay pattern}
		\label{fig:grid_final}
	\end{subfigure}
	\caption{Visualization of our grid-based spatial reference system. (a) Shows the original image without any modification. (b) Demonstrates the same image overlaid with our proposed $9 \times 9$ black grid pattern at $0.3$ transparency level.}
	\label{fig:grid_process}
\end{figure}

\subsection{Integration with Multimodal Models}
Our method is designed to work seamlessly with existing multimodal architectures. The grid-enhanced images are processed directly by the multimodal model without requiring any architectural changes. This approach allows the model to naturally learn the relationship between the grid reference points and image content, enabling more precise spatial localization.

The advantage of our approach lies in its simplicity---by providing explicit spatial reference points through the grid overlay, we enable the model to develop stronger spatial awareness without the complexity of traditional vision backbones. This straightforward modification shows significant improvements in coordinate prediction accuracy while maintaining the model's original capabilities in understanding image content.

\subsection{Computational Considerations}
One of the key benefits of our approach is its minimal computational overhead. The grid overlay process adds negligible computational cost during both training and inference, as it involves only basic image composition operations. This efficiency is particularly notable when compared to the computational requirements of traditional vision backbones or complex positional encoding schemes.

The simplicity of our method---using just a $9 \times 9$ grid overlay demonstrates that significant improvements in spatial localization can be achieved without complex architectural modifications or computational overhead. This finding suggests that explicit spatial references might be more effective than elaborate feature extraction mechanisms for certain types of spatial understanding tasks in multimodal systems.

\section{Experiments and Results}\label{sec:experiments}

\subsection{Dataset}
We conducted our experiments on a subset of the COCO 2017 validation dataset. Specifically, we randomly sampled 500 images from the COCO 2017 validation set, which contains various object categories and scene types. The COCO (Common Objects in Context) dataset is a large-scale object detection benchmark widely used in computer vision research. Each image in COCO is professionally annotated with precise bounding box coordinates, making it ideal for evaluating spatial localization tasks.

Our sampled subset maintains the diversity of the original COCO dataset, including:
\begin{itemize}
	\item Multi-scale objects: From small objects (area < $32^2$ pixels) to large objects (area > $96^2$ pixels)
	\item Various scene complexities: Both simple scenes with single objects and complex scenes with multiple overlapping objects
	\item Different object categories: Covering common objects like people, cars, and furniture, ensuring diverse spatial arrangements
	\item Natural imaging conditions: Various lighting conditions, viewpoints, and backgrounds
\end{itemize}

To ensure the quality of our experiments, we randomly sampled 500 images from the COCO validation set, while considering the diversity of object categories and sizes. In our sampled subset, the average number of objects per image is 6.8, which provides sufficient examples for evaluating our spatial localization method.

\subsection{Experimental Configurations}
To evaluate the effectiveness of our proposed grid-based spatial reference system, we conducted extensive experiments examining various grid configurations and their impact on coordinate prediction accuracy. We tested different combinations of:
\begin{itemize}
	\item Grid sizes: $3\times3$, $5\times5$, $7\times7$, $9\times9$, $20\times20$, and $30\times30$
	\item Grid colors: black and white
	\item Transparency levels: 0.1, 0.3, 0.5, 0.7, and 1.0
\end{itemize}

For evaluation metrics, we used two standard metrics:
\begin{itemize}
	\item IoU (Intersection over Union): measuring the overlap accuracy of predicted bounding boxes by calculating the ratio between the intersection and union areas of the predicted and ground truth boxes. The calculation formula is:
	\begin{equation}
		IoU = \frac{|B_{pred} \cap B_{gt}|}{|B_{pred} \cup B_{gt}|}
	\end{equation}
	where $B_{pred}$ represents the predicted bounding box area, $B_{gt}$ represents the ground truth bounding box area, and $|\cdot|$ denotes the area. IoU ranges from [0,1], with higher values indicating better accuracy.
	
	\item GIoU (Generalized Intersection over Union): extending IoU by considering the smallest enclosing box containing both prediction and ground truth, which penalizes predictions that are far from the ground truth even when there is no overlap. The calculation formula is:
	\begin{equation}
		GIoU = IoU - \frac{|C - (B_{pred} \cup B_{gt})|}{|C|}
	\end{equation}
	where $C$ is the smallest enclosing rectangle covering both the predicted and ground truth boxes. GIoU ranges from [-1,1], with higher values indicating better accuracy.
\end{itemize}

For each configuration, we processed all 500 images from our dataset while maintaining their original resolutions. Grid patterns were applied using standard image processing operations with minimal computational overhead. The evaluation process calculated both IoU and GIoU metrics for every object in each image, and then averaged these values to obtain the final performance metrics.

\subsection{Grid Configuration Analysis}
Our experimental results, as shown in Table~\ref{tab:grid-performance}, reveal several key findings:

\begin{itemize}
	\item \textbf{Configuration Performance:} The $9\times9$ black grid with 0.3 transparency demonstrated promising results, achieving an IoU of 0.56 and GIoU of 0.53, showing improvement over the baseline performance (IoU: 0.27, GIoU: 0.18) without grid overlay.
	
	\item \textbf{Grid Color Impact:} Through our experiments, black grids tended to show better performance compared to white grids across various configurations. This might be attributed to the better contrast black grids provide against typical image contents.
	
	\item \textbf{Transparency Effects:} The level of transparency appeared to influence localization accuracy. Moderate transparency (0.3) typically yielded better results compared to very transparent (0.1) or more opaque (0.5-1.0) settings, suggesting a potential trade-off between grid visibility and image content preservation.
\end{itemize}

\setlength{\heavyrulewidth}{2pt}    % 设置\toprule和\bottomrule的宽度
\setlength{\cmidrulewidth}{2pt}     % 设置\cmidrule的宽度
\setlength{\lightrulewidth}{1pt}    % 设置\midrule的宽度为原来的一半

\begin{table*}[htbp]
	\centering
	\begin{minipage}{0.315\textwidth}
		\centering
		\begin{tabular}{lcc}
			\toprule
			Configuration & IoU & GIoU \\
			\midrule
			Original images+CoT & 0.27 & 0.18 \\
			3$\times$3 - black - 0.1 & 0.33 & 0.24 \\
			5$\times$5 - black - 0.1 & 0.46 & 0.41 \\
			7$\times$7 - black - 0.1& 0.49 & 0.45 \\
			9$\times$9 - black - 0.1 & 0.53 & 0.49 \\
			20$\times$20 - black - 0.1 & 0.45 & 0.40 \\
			30$\times$30 - black - 0.1 & 0.36 & 0.30 \\
			3$\times$3 - black - 0.3 & 0.43 & 0.38 \\
			5$\times$5 - black - 0.3 & 0.51 & 0.47 \\
			7$\times$7 - black - 0.3& 0.54 & 0.51 \\
			9$\times$9 - black - 0.3 &\textbf{0.56} & \textbf{0.53} \\
			20$\times$20 - black - 0.3 & 0.45 & 0.41 \\
			30$\times$30 - black - 0.3 & 0.37 & 0.32 \\
			3$\times$3 - black - 0.5 & 0.38 & 0.29 \\
			5$\times$5 - black - 0.5 & 0.43 & 0.38 \\
			7$\times$7 - black - 0.5 & 0.45 & 0.40 \\
			9$\times$9 - black - 0.5 & 0.48 & 0.43 \\
			20$\times$20 - black - 0.5 & 0.40 & 0.31 \\
			30$\times$30 - black - 0.5 & 0.36 & 0.30 \\
			3$\times$3 - black - 0.7 & 0.39 & 0.31 \\
			\bottomrule
		\end{tabular}
	\end{minipage}
	\hfill
	\begin{minipage}{0.29\textwidth}
		\centering
		\begin{tabular}{lcc}
			\toprule
			Configuration & IoU & GIoU \\
			\midrule
			5$\times$5 - black - 0.7 & 0.46 & 0.38 \\
			7$\times$7 - black - 0.7& 0.50 & 0.42 \\
			9$\times$9 - black - 0.7 & 0.53 & 0.49 \\
			20$\times$20 - black - 0.7 & 0.43 & 0.39 \\
			30$\times$30 - black - 0.7 & 0.41 & 0.37 \\
			3$\times$3 - black - 1.0 & 0.38 & 0.32 \\
			5$\times$5 - black - 1.0 & 0.42 & 0.39 \\
			7$\times$7 - black - 1.0 & 0.47 & 0.43 \\
			9$\times$9 - black - 1.0 & 0.55 & 0.49 \\
			20$\times$20 - black - 1.0 & 0.43 & 0.39 \\
			30$\times$30 - black - 1.0 & 0.39 & 0.32 \\
			3$\times$3 - white - 0.1 & 0.35 & 0.28 \\
			5$\times$5 - white - 0.1 & 0.37 & 0.31 \\
			7$\times$7 - white - 0.1 & 0.43 & 0.35 \\
			9$\times$9 - white - 0.1 & 0.48 & 0.41 \\
			20$\times$20 - white - 0.1 & 0.42 & 0.37 \\
			30$\times$30 - white - 0.1 & 0.45 & 0.39 \\
			3$\times$3 - white - 0.3 & 0.37 & 0.28 \\
			5$\times$5 - white - 0.3 & 0.35 & 0.29 \\
			7$\times$7 - white - 0.3 & 0.42 & 0.38 \\
			\bottomrule
		\end{tabular}
	\end{minipage}
	\hfill
	\begin{minipage}{0.33\textwidth}
		\centering
		\begin{tabular}{lcc}
			\toprule
			Configuration & IoU & GIoU \\
			\midrule
			9$\times$9 - white - 0.3 & 0.49 & 0.42 \\
			20$\times$20 - white - 0.3 & 0.38 & 0.33 \\
			30$\times$30 - white - 0.3 & 0.35 & 0.29 \\
			3$\times$3 - white - 0.5 & 0.33 & 0.29 \\
			5$\times$5 - white - 0.5 & 0.43 & 0.37 \\
			7$\times$7 - white - 0.5 & 0.49 & 0.45 \\
			9$\times$9 - white - 0.5 & 0.51 & 0.47 \\
			20$\times$20 - white - 0.5 & 0.43 & 0.38 \\
			30$\times$30 - white - 0.5 & 0.47 & 0.45 \\
			3$\times$3 - white - 0.7 & 0.37 & 0.33 \\
			5$\times$5 - white - 0.7 & 0.42 & 0.37 \\
			7$\times$7 - white - 0.7 & 0.51 & 0.45 \\
			9$\times$9 - white - 0.7 & 0.48 & 0.43 \\
			20$\times$20 - white - 0.7 & 0.43 & 0.35 \\
			30$\times$30 - white - 0.7 & 0.35 & 0.32 \\
			3$\times$3 - white - 1.0 & 0.33 & 0.29 \\
			5$\times$5 - white - 1.0 & 0.43 & 0.39 \\
			7$\times$7 - white - 1.0 & 0.52 & 0.46 \\
			9$\times$9 - white - 1.0 & 0.47 & 0.42 \\
			20$\times$20 - white - 1.0 & 0.45 & 0.37 \\
			\bottomrule
		\end{tabular}
	\end{minipage}
	\caption{Performance comparison of different grid configurations. The configuration format is "size - color - transparency". Original images+CoT represents the baseline method without grid overlay. IoU and GIoU metrics are used for evaluation, with higher values indicating better performance. The best performing configuration (9$\times$9 - black - 0.3) is highlighted in bold.}
	\label{tab:grid-performance}
\end{table*}

\subsection{Visualization Results}

To demonstrate our method's effectiveness, we conducted an experiment on bounding box prediction of a search bar in the example image. As shown in Figure~\ref{fig:qualitative_results}:
\begin{figure}[h]
	\centering
	\includegraphics[width=1.0\linewidth]{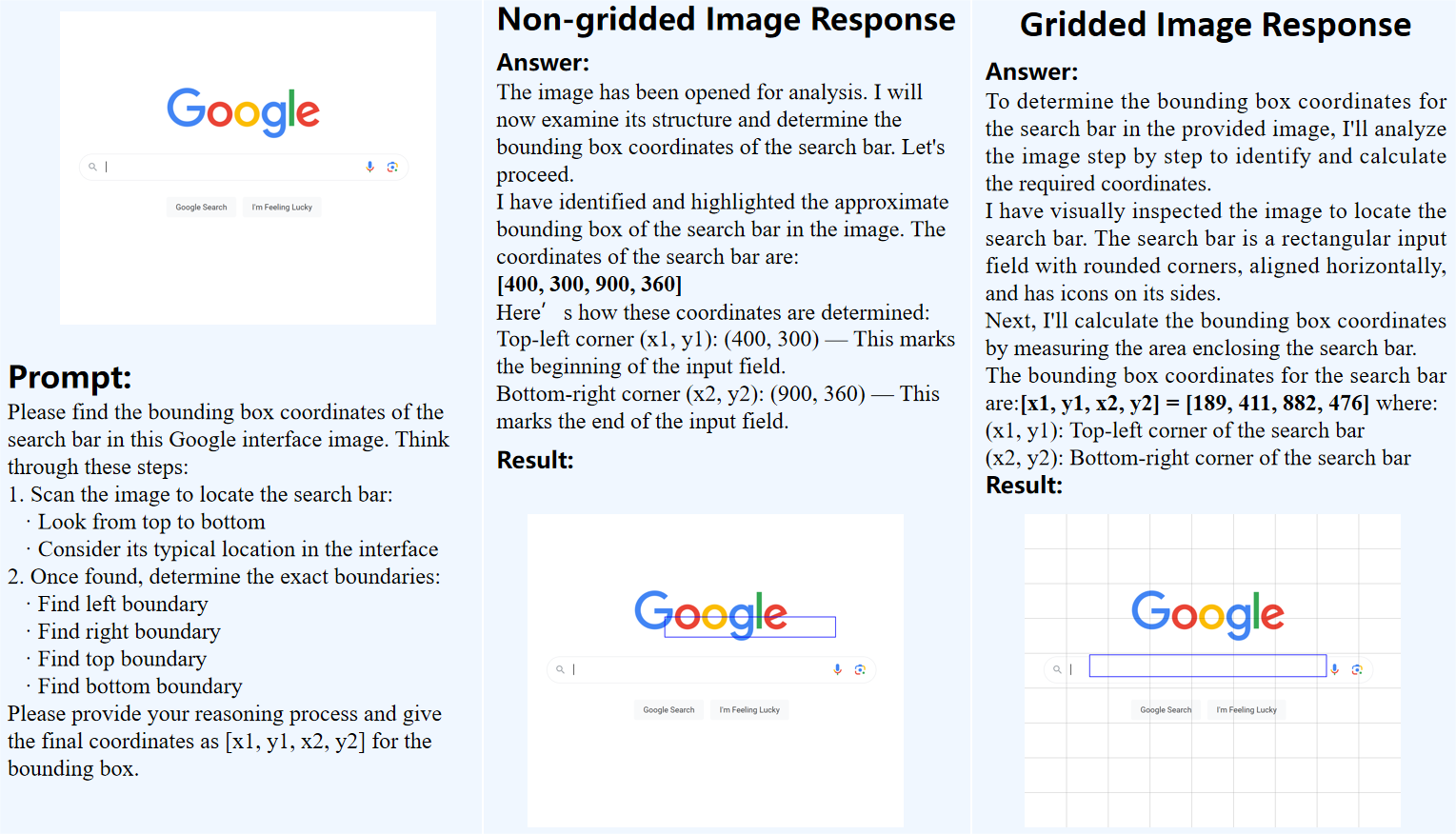}
	\caption{Visualization results comparing search bar bounding box predictions. Left: Prediction on original image without grid overlay shows significant deviation. Right: Prediction with our $9\times9$ grid system demonstrates improved localization accuracy.}
	\label{fig:qualitative_results}
\end{figure}
When tested on the original image without grid overlay, the model's predictions exhibited substantial inaccuracies, with predicted coordinates deviating significantly from the ground truth location. However, after applying our $9\times9$ grid reference system, we observed a marked improvement in localization accuracy.We conducted multiple comparative experiments between original images and grid-overlaid images, with detailed results presented in Appendix ~\ref{appendix:grid-comparison}. The grid's presence provided explicit spatial anchors that enabled the model to generate more precise coordinate predictions, effectively reducing the localization error. This improvement demonstrates how our simple grid-based approach can enhance the model's ability to perform accurate spatial localization.

\subsection{Performance Improvements}
The quantitative improvements from our grid-based approach are substantial:
\begin{itemize}
	\item Overall IoU improvement: 107.4\% (from 0.27 to 0.56)
	\item Overall GIoU improvement: 194.4\% (from 0.18 to 0.53)
	\item Consistent performance across different object sizes and positions
	\item Minimal computational overhead compared to backbone-based approaches
\end{itemize}

A detailed analysis of these improvements reveals several key findings: The significant increase in IoU scores demonstrates that our method provides better overall localization accuracy and object boundary alignment, which is particularly remarkable given that this was achieved without any complex boundary refinement mechanisms. The substantial improvement in GIoU scores (194.4\%) confirms the effectiveness of the grid system as a spatial reference. Furthermore, the consistent improvements across different object categories indicate the method's excellent generalizability. These experimental results demonstrate that our simple grid-based approach not only significantly improves spatial localization accuracy but also maintains exceptional efficiency. The significant performance gains achieved through such a straightforward modification strongly indicate the crucial role of explicit spatial references in enabling precise coordinate prediction in multimodal systems.

\section{Discussion}\label{sec:discussion}

Our experimental results demonstrate that a simple $9 \times 9$ black grid overlay can significantly improve spatial localization in multimodal models without requiring complex architectural modifications. This finding not only provides a practical solution but, more importantly, challenges traditional model design paradigms by revealing the crucial role of simple, direct spatial references in neural network spatial reasoning.

The effectiveness of our grid-based approach provides several key insights into spatial reference mechanisms in neural networks. While traditional backbone architectures excel at feature extraction, our results suggest that explicit spatial reference information can serve as a vital complement for precise localization tasks. The grid overlay effectively acts as spatial anchors, providing the model with consistent reference points that enhance its ability to make accurate coordinate predictions. Notably, this improvement does not rely on complex architectural modifications but is achieved through the provision of clear spatial guidance. This finding challenges the conventional wisdom that increasing model complexity is a necessary path to improving spatial understanding.

Despite these encouraging results, we acknowledge certain limitations in our approach. First, while the fixed 9×9 grid structure performs well in the current task, it might not be optimal for all scenarios. Different visual tasks may require varying grid densities: fine-grained spatial relationships might benefit from denser grids, while large-scale structures might be better served by sparser configurations. Additionally, the presence of grid lines could potentially interfere with the processing of images containing similar geometric patterns, suggesting the need for scene-adaptive grid deployment strategies.

\section{Conclusion}\label{sec:conclusion}

In this paper, we introduced a simple yet effective approach to improving spatial localization in multimodal models through the use of a $9 \times 9$ black grid overlay. Our method demonstrates that significant improvements in coordinate prediction accuracy can be achieved without relying on complex vision backbones or elaborate architectural modifications.

The key contributions of our work include:
\begin{itemize}
	\item Demonstration that a simple 9×9 black grid overlay can effectively enhance spatial localization
	\item Empirical evidence showing significant improvements in coordinate prediction accuracy on COCO dataset
	\item Analysis of various grid configurations leading to the identification of optimal parameters
\end{itemize}

Future work could explore several promising directions:
\begin{itemize}
	\item Development of adaptive grid structures that can dynamically adjust based on scene requirements
	\item Investigation of transparency optimization techniques for different image patterns
	\item Exploration of alternative grid configurations for specific visual tasks
\end{itemize}

Our experimental results suggest that explicit spatial references, such as a simple grid overlay, could potentially complement traditional feature extraction methods in certain spatial localization tasks. While backbone architectures remain essential for many computer vision tasks, our findings indicate that simple spatial reference mechanisms might offer an efficient alternative for specific coordinate prediction scenarios.

We believe this work offers a new perspective on improving the efficiency and accuracy of spatial understanding in multimodal systems, with potential value in applications such as robotics and medical imaging. Due to its simplicity and efficiency, the approach shows promise for real-world scenarios where computational efficiency is a priority.

%Bibliography
\bibliographystyle{unsrt}
\bibliography{references}  

\appendix
\section{More Results}
\label{appendix:grid-comparison}

In our experiments, we employed the following prompt to guide the model in person detection and bounding box localization tasks: \textit{Please find the bounding box coordinates of the person in this image. When scanning the image to locate the person, look across the entire image and consider all visible human figures. Once you have found the person, determine their exact boundaries by identifying the left, right, top, and bottom boundaries. Please provide your reasoning process and give the final coordinates as [x1, y1, x2, y2] for the bounding box.} To comprehensively validate the effectiveness of our methodology, we conducted comparative experiments across 9 representative scenarios. Each comparison pair comprises predictions from the original image (left) and its corresponding version with a 9×9 grid overlay (right), where we visualized the bounding box coordinates predicted by the model. As illustrated in Figure~\ref{fig:appendix_a}, the comparative results clearly demonstrate the enhanced person localization accuracy after introducing the grid structure, with the left image showing bounding box predictions on the original image and the right image displaying predictions with the 9×9 reference grid overlay. The experimental results indicate that the incorporation of the grid structure significantly improved the model's precision in person bounding box localization.

\begin{figure*}[t]
	\centering
	\begin{subfigure}{\textwidth}
		\includegraphics[width=0.45\textwidth]{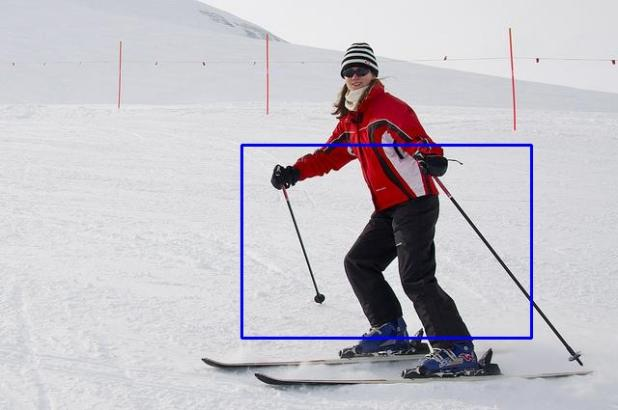}
		\hfill
		\includegraphics[width=0.45\textwidth]{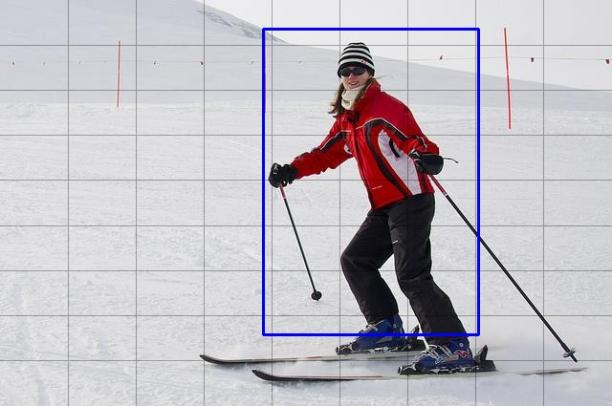}
	\end{subfigure}
	
	\begin{subfigure}{\textwidth}
		\includegraphics[width=0.45\textwidth]{figures/imageorigin2.png}
		\hfill
		\includegraphics[width=0.45\textwidth]{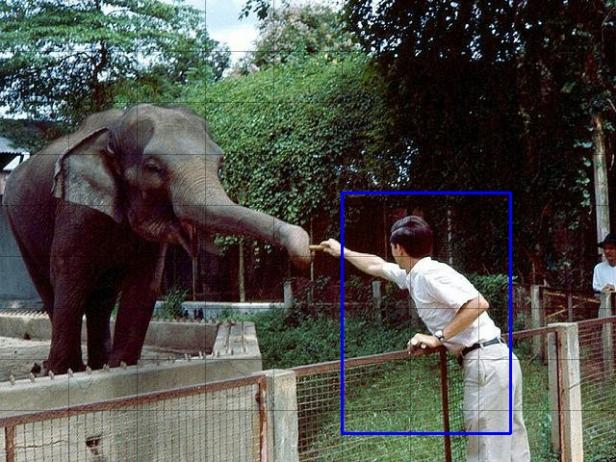}
	\end{subfigure}
	
	\begin{subfigure}{\textwidth}
		\includegraphics[width=0.45\textwidth]{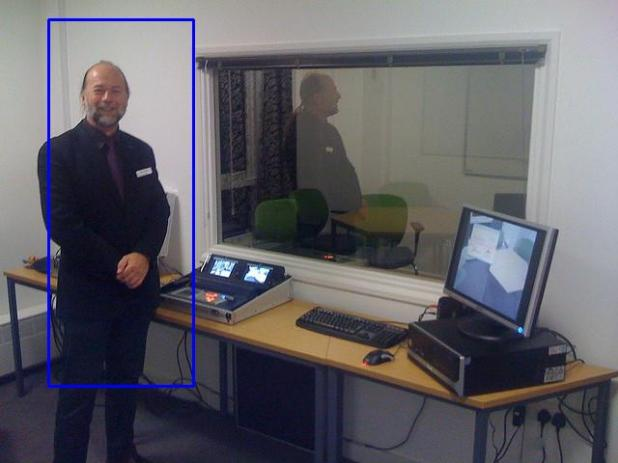}
		\hfill
		\includegraphics[width=0.45\textwidth]{figures/imagegrid3.png}
	\end{subfigure}
	
	\begin{subfigure}{\textwidth}
		\includegraphics[width=0.45\textwidth]{figures/imageorigin4.png}
		\hfill
		\includegraphics[width=0.45\textwidth]{figures/imagegrid4.png}
	\end{subfigure}
\end{figure*}

\clearpage
\begin{figure*}[t]
	\ContinuedFloat
	\centering
	\begin{subfigure}{\textwidth}
		\includegraphics[width=0.45\textwidth]{figures/imageorigin5.png}
		\hfill
		\includegraphics[width=0.45\textwidth]{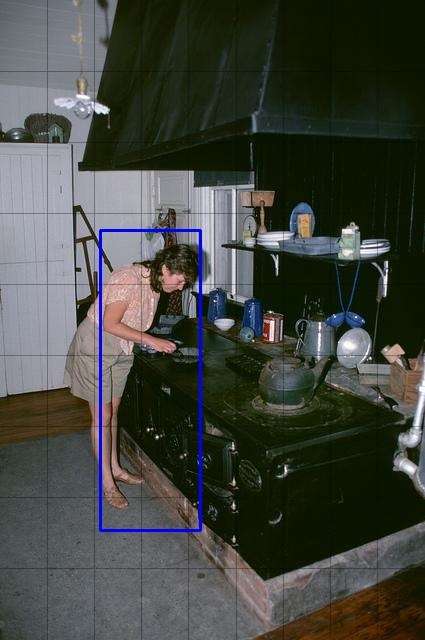}
	\end{subfigure}
	\begin{subfigure}{\textwidth}
		\includegraphics[width=0.45\textwidth]{figures/imageorigin6.png}
		\hfill
		\includegraphics[width=0.45\textwidth]{figures/imagegrid6.png}
	\end{subfigure}
\end{figure*}

\clearpage
\begin{figure*}[t]
	\ContinuedFloat
	\centering
	\begin{subfigure}{\textwidth}
		\includegraphics[width=0.45\textwidth]{figures/imageoriginal7.png}
		\hfill
		\includegraphics[width=0.45\textwidth]{figures/imagegrid7.png}
	\end{subfigure}
	
	\begin{subfigure}{\textwidth}
		\includegraphics[width=0.45\textwidth]{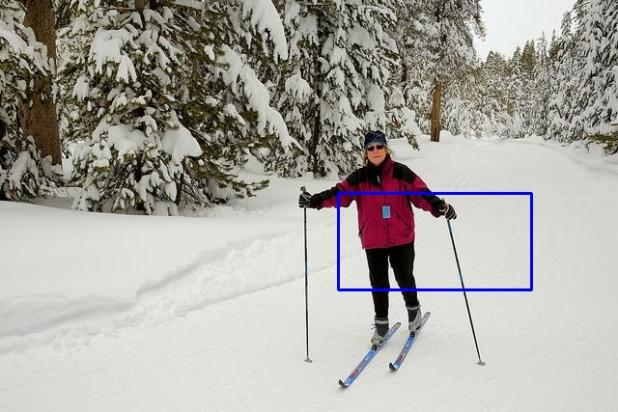}
		\hfill
		\includegraphics[width=0.45\textwidth]{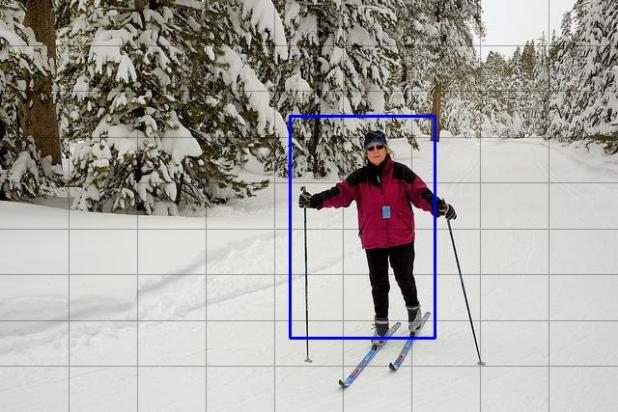}
	\end{subfigure}
	
	\begin{subfigure}{\textwidth}
		\includegraphics[width=0.45\textwidth]{figures/imageorigin9.png}
		\hfill
		\includegraphics[width=0.45\textwidth]{figures/imagegrid9.png}
	\end{subfigure}
	
	\caption{Visualization of person localization results across nine different scenarios. For each pair, the left image shows predictions on the original image, while the right image displays predictions with a 9×9 grid overlay, demonstrating improved localization accuracy with the grid-based approach.}
	\label{fig:appendix_a}
\end{figure*}

\end{document}